\documentclass[10pt,twocolumn,letterpaper]{article}

\usepackage{iccv}
\usepackage{times}
\usepackage{epsfig}
\usepackage{graphicx}
\usepackage{amsmath}
\usepackage{amssymb}

\usepackage{comment}
\usepackage{subcaption,caption}

\usepackage[pagebackref=true,breaklinks=true,letterpaper=true,colorlinks,bookmarks=false]{hyperref}

 \iccvfinalcopy 

\def\iccvPaperID{1693} 

\ificcvfinal\fi
    \definecolor{MyDarkBlue}{rgb}{0,0.08,0.5}
    \definecolor{MyDarkGreen}{rgb}{0.02,0.30,0.02}
    \definecolor{MyDarkRed}{rgb}{0.7,0.02,0.02}
    \definecolor{MyDarkOrange}{rgb}{0.40,0.2,0.02}

\newcommand{\red}[1]{\textcolor{red}{#1}}

\begin{document}

\title{Understanding Intra-Class Knowledge Inside CNN}
\author{
Donglai Wei \
Bolei Zhou \
Antonio Torrabla \
William Freeman\\
\{donglai, bzhou, torrabla, billf\} @csail.mit.edu
}

\maketitle

\begin{abstract}
Convolutional Neural Network (CNN)
has been successful in image recognition tasks,
and recent works shed lights on how CNN separates different classes with the learned inter-class knowledge through visualization~\cite{Mahendran15,Simonyan14,Zhou14}. 
In this work, we instead visualize the intra-class knowledge inside CNN to better understand how an object class is represented in the fully-connected layers.
To invert the intra-class knowledge into more interpretable images,
we propose a non-parametric patch prior upon previous CNN visualization models~\cite{Mahendran15,Simonyan14}.
With it, we show how different ``styles" of templates for an object class are organized by CNN in terms of location and content, and represented in a hierarchical and ensemble way.
Moreover, such intra-class knowledge can be used in many interesting applications, e.g. style-based image retrieval and style-based object completion.
\end{abstract}

\section{Introduction}
Deep Convolutional neural networks (CNN)~\cite{Lecun98} achieve the state-of-the-art performance at recognition tasks. 
Recent works~\cite{Zeiler14,Agrawal14,Szegedy13} have focused on understanding the inter-class discriminative power of CNN. In particular, ~\cite{Zhou14} shows that 
individual neurons in different convolutional layers
correspond to texture patterns with various level of abstraction and even object detectors can be found in the last feature extraction layer.

\begin{figure}[t]
\centering
\includegraphics[width=0.5\textwidth]{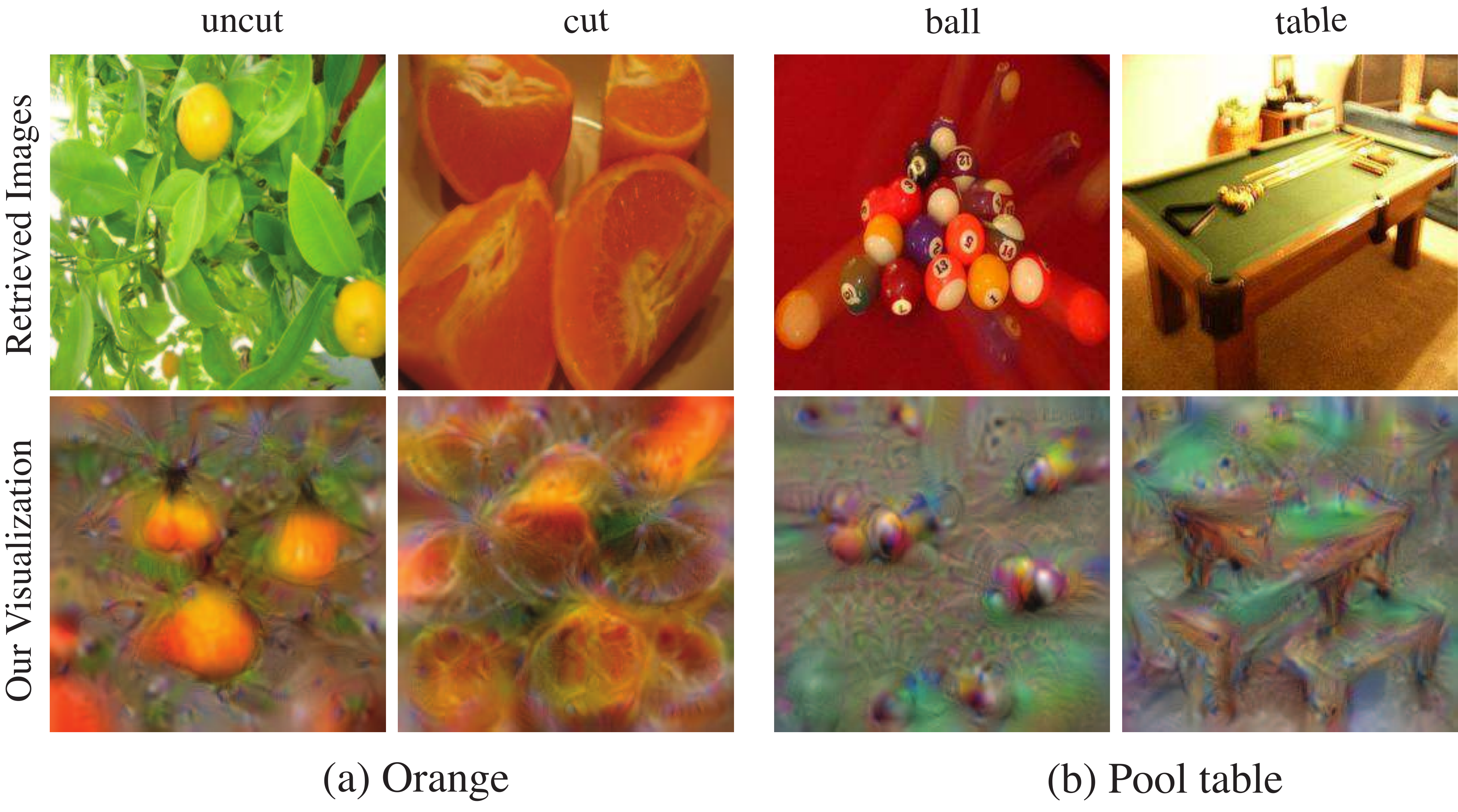}
\caption{Examples of intra-class variation. 
We show two different styles of the object class ``orange" and ``pool table'' with the retrieved images and our new visualization method.}
\label{fig:teaser}
\end{figure}

However, little is known about how CNN represent an object class or how it captures the intra-class variation.
For example, in the object class of ``orange'' and ``pool table'', there are drastically different ``styles'' of the object instances which CNN recognizes correctly (Fig.~\ref{fig:teaser}).
There are two main challenges of this problem.
One is to visualize the knowledge numerically instead of directly retrieving natural images, which can be biased towards the image database that is in use.
The other challenge is that such intra-class knowledge is captured collectively by a group of neurons, namely ``neural pathway'', instead of a single neuron studied in previous works.

In this work, we make progress on both challenges by 
(1) introducing a patch prior to improve parametric CNN visualization models,
(2) analyzing how the spatial and style intra-class knowledge are encoded inside CNN in a hierarchical and ensemble way.
With this learned knowledge, we can retrieve images or complete images in a novel way.
Our techniques apply to a range of feedfoward architectures 
and we here focus on the CNN~\cite{Krizhevsky12} trained on the large-scale ImageNet challenge dataset~\cite{Deng09},
with 5 convolutional layers
followed by 3 fully-connected layers.

\section{Related Work}
Below, we survey works on understanding fully-connected layers and intra-class knowledge discovery.

\subsection{Fully-Connected Layers in CNN}
\noindent\textbf{Understanding}
Below are some recent understandings of fully-connected layers.
(1) Dropout Techinques. 
~\cite{Krizhevsky12} consider the dropout technique as an approximation of learning ensemble models and  ~\cite{Baldi13} proves its equivalence to a regularization;
(2) Binary Code.
~\cite{Agrawal14} discovers that the biniary mask of the features from fc$_{6-7}$ layers are good enough for classification. 
(3) Pool5.
$p_5$ features contain object parts information with spatial and semantic. we can combine them by selecting sub-matrices in $W_6$
(4) Image Retrival from fc$_7$:
fc$_{7}$ is used as semantic space

\noindent\textbf{Visualization}
Unlike features in convolutional layers
where we can recover most of the original images with parametric~\cite{Zeiler14,Mahendran15} or non-parametric methods,
features from fully-connected are hard to invert.
As shown in~\cite{Mahendran15},
the location and style information of the object parts are lost.
Another work~\cite{Simonyan14} inverts the class-specific feature from fc$_8$ layer which is 0 except the target class. 
The output image from numerical optimization is a composite of various object templates.
Both these works follow the same model framework (compared in Sec.~\ref{subsec:model_inv_pm}) 
which can be solved efficiently with gradient descend method. 

\subsection{Intra-class Knowledge Discovery}
Understanding image collections is a relatively unexplored task, although there is growing interest in this area. Several methods attempt to represent the continuous variaation in an image class using sub-spaces or manifolds. Unlike this work, we investigate discrete, name- able transformations, like crinkling, rather than working in a hard-to-interpret parameter space. Photo collections have also been mined for storylines as well as spatial and temporal trends, and systems have been proposed for more general knowledge discovery from big visual data. ~\cite{Isola15} focuses on physical state transformations, and in addition to discovering states it also studies state pairs that define a transformation.

In Sec.~\ref{sec:model_conv}, we analyze the problem of current parametric CNN visualization models and propose a data-driven patch prior to generate images with natural color distribution.
In Sec.~\ref{sec:model_fc}, we decompose the fully-connected layers into four different components, which are shown to capture the the location-specific and content-specific intra-class variation, or represent such knowledge in a hierarchical and ensemble way.
In Sec.~\ref{sec:result}, we first provide both quantitative and qualitative results for our new visualization methods.
We apply the learned intra-class knowledge inside CNN to organize an unlabelled image collection and to fill in image masks with objects of various styles.

\begin{figure}[ht]
\centering
\includegraphics[width=0.5\textwidth]{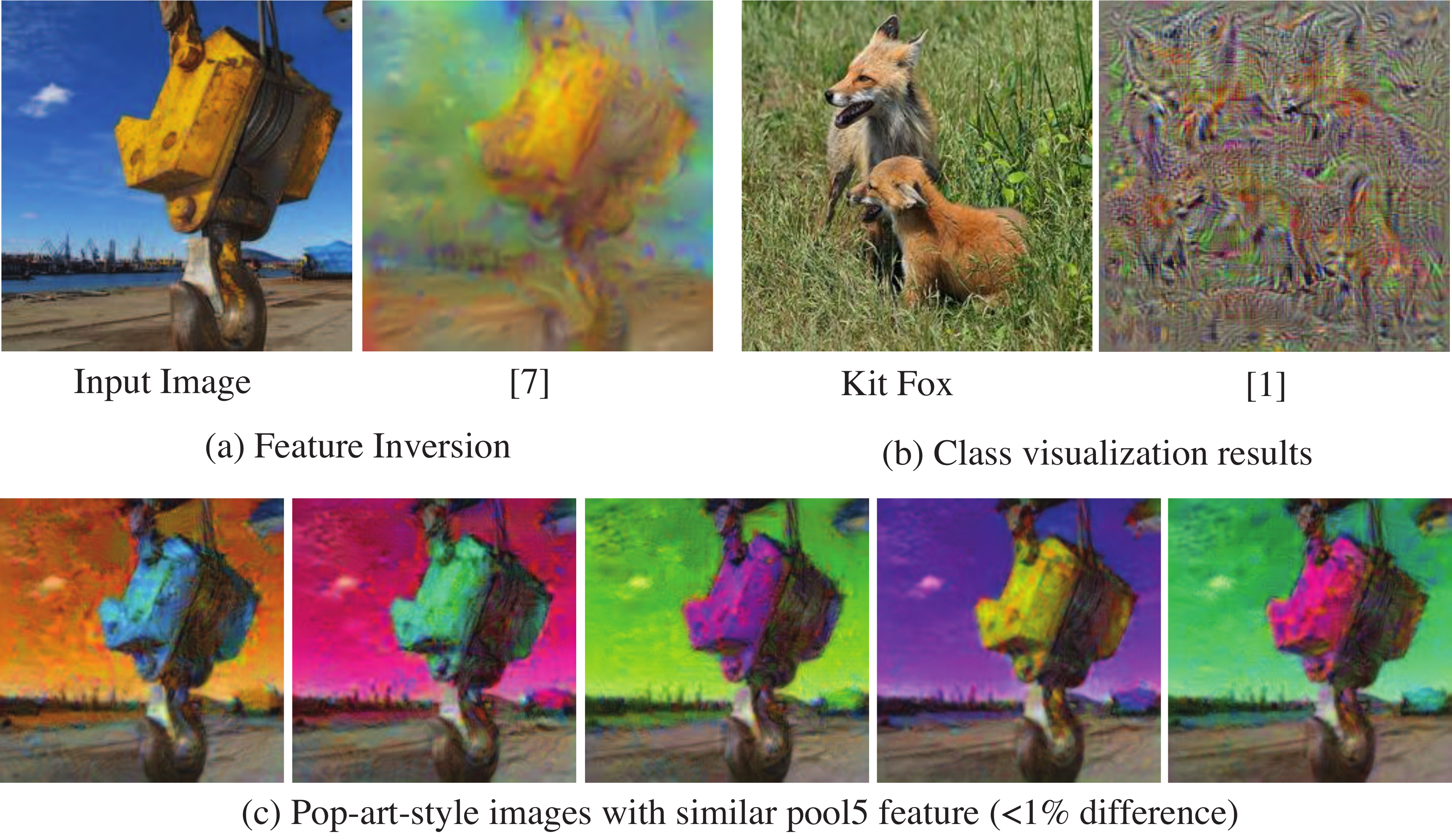}
\caption{Illustration of the color problem in~\cite{Simonyan14}. The results for (a) pool$_5$ feature inversion and (b) class visualization have unnatural global color distribution. In (c), we show six output images with different global color distribution, but similar pool5 features differed less than 1\% from each other.}
\label{fig:inv_p5}
\end{figure}
\section{CNN Visualization with Patch Prior}
~\label{sec:model_conv}
\begin{figure}[t]
\centering
\includegraphics[width=0.5\textwidth]{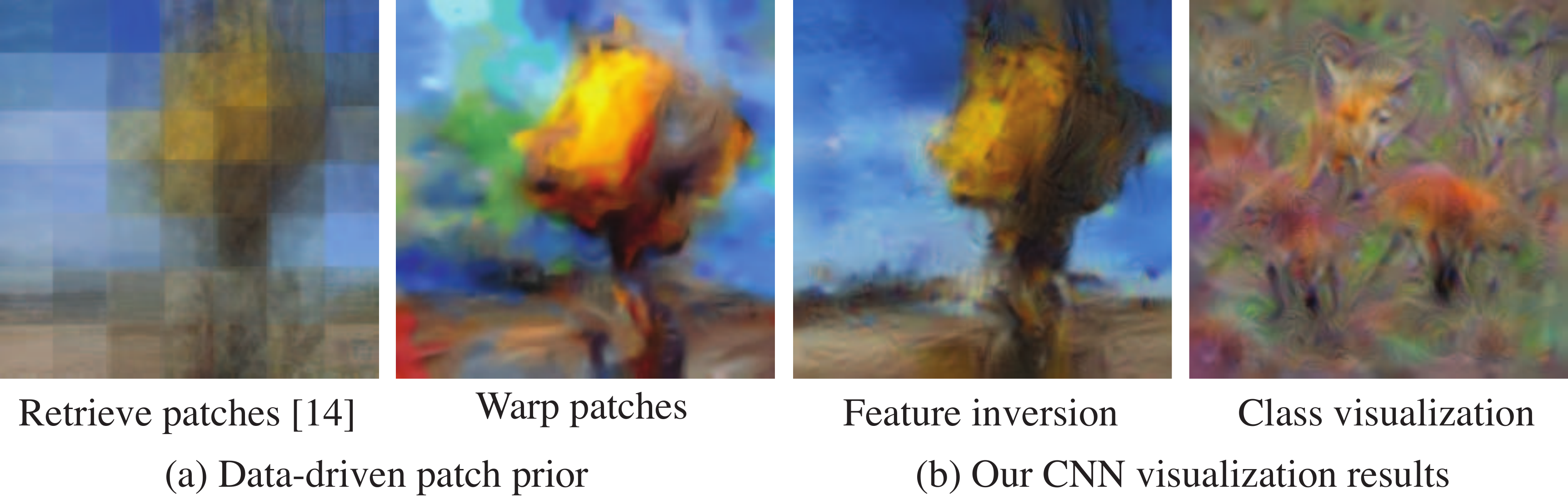}
\caption{Illustration of the local optima of pool5 feature inversion. We show six output images with different global color distribution, but similar pool5 features differed less than 1\% from each other.}
\label{fig:inv_p5_ours}
\end{figure}

Below, we propose a data-driven patch prior to improve parametric CNN visualization models~\cite{Mahendran15,Simonyan14} 
and we show improvement for both cases (Fig.~\ref{fig:inv_p5_ours}\red{b}).

\subsection{Parametric Visualization Model}
~\label{subsec:model_inv_pm}
We first consider the task of feature inversion~\cite{Mahendran15}.
Given the CNN feature (e.g. pool$_5$) of a natural image,
the goal is to invert it back to an image close to the original.
~\cite{Mahendran15} aims to find an optimal image 
that minimizes the sum of the data energy from feature reconstruction error 
and a regularization energy $R(I)$ for the estimation.
\begin{align}
E^{k}_{inv}(I)&= \frac{\|\Phi_k(I)-\phi_0)\|_2^2}{\|\phi_0\|^2_2} + R(I),
\label{Eqn:E_inv}
\end{align}
where $\Phi_k$ is the CNN feature from layer $k$,
$\phi_0$ is the target feature for inversion. 
Similarly, another CNN visualization task, class visualization~\cite{Simonyan14},
follows a similar formulation,
where the goal is to generate an image given the class label $t$.
\begin{align}
E_{class}(I)&=\Phi_8^t\cdot\Phi_8(I)+R(I),\label{Eqn:E_f8}
\end{align}
where $\Phi_8^t$ is the binary vector with only the $t$-th element one.

For the regularization term $R(I)$,
the $\alpha$-norm of the image $\|I\|_{\alpha}^{\alpha}$~\cite{Simonyan14} and the pairwise gradient$\|\nabla I\|_{\beta}^{\beta}$~\cite{Mahendran15} are used. 
Unlike low-level vision reconstruction (e.g. denoising), the data energy from CNN is less sensitive to low-frequency image content, 
which leads to multiple global optima with unnatural color distribution.
Given the input image (Fig.~\ref{fig:inv_p5}\red{a}),
we show a collection of pop-art style images whose pool$_5$ features are less than 1\% from the input
(Fig.~\ref{fig:inv_p5}\red{c}).
These images are generated from~\cite{Mahendran15}, initialized from the input image with shuffled RGB channels. 
In practice, ~\cite{Simonyan14,Mahendran15} initialize the optimization from the mean image with or without white noise,
and the gradient descend algorithm converges to one of the global optima
whose color distribution can be far from being natural (Fig.~\ref{fig:inv_p5}\red{a-b}).

\subsection{Data-driven Patch Prior}
~\label{subsec:model_inv_pp}
To regularize the color distribution for CNN visualization, we build an external database of natural patches 
and minimize the distance of patches from the output to those in the database.
As the patches from the CNN visualization models above are lack of low-frequency components, we calculate the distance between patches after global normalization w.r.t the mean and std of the whole image respectively. 
Combined with previous regularization models, our final image regularization model is
\begin{align}
R(I)=R_{\alpha}\|I\|_2^2+R_{\beta}\|\nabla I\|_2^2+R_{\gamma}\sum_p\|\tilde{I_p}- \tilde{D_p}\|_2^2,
\label{Eqn:inv_p5_r}
\end{align}
where $R_{\alpha-\gamma}$ are weight parameters for each term, $p$ is the patch index, $\tilde{I_p}$ are the densely sampled normalized patches and $D_p$ are the nearest normalized patches from a natural patch database. 
In practice, we iteratively solve the continuous optimizaiton for $I$ given the matched patches $D_p$
and the discrete optimization for $D_p$ with patch match given the previous estimate of $I$.

To illustrate the effectiveness of the patch prior, we 
compute the dense patch correspondence from the patch database to a pool$_5$ feature inversion result (Fig.~\ref{fig:inv_p5_ours}\red{a}) ~\cite{Mahendran15}, 
and visualize the warped image which regularizes the output image in Eqn.~\ref{Eqn:inv_p5_r}.
We compare the patch matching quality with and without normalization. 
As expected, the normalized patches have better chance to retrieve natural patches, and the warped result is reasonable despite the unnatural color distribution of the initial estimation.

Below, we describe how to build an effective patch database.
The object class visualization task has no ground truth color distribution and we can directly sample patches from validation images from the same class.
For feature inversion, however, such approach can be costly due to the intra-class variation of each object class, where images from the same class may not match well.
As discovered in~\cite{Long14},
conv-layer features can be used to retrieve image patches with similar appearance, 
though their semantic can be totally different.
Thus, we build a database of 1M pairs of 1x1x256 pool5 features and the center 67x67x3 patch of the original patch support (195x195x3). 
Given the pool$_5$ feature to invert, we build our patch database with such retrieval methods and 
we show the averaged patches (10-NN at each pool$_5$ location) recovers well the color distribution of the input image
 (Fig.~\ref{fig:inv_p5_ours}\red{a}).

 \section{Discover CNN Intra-class Knowledge}
~\label{sec:model_fc}
\begin{figure}[t]
\centering
\includegraphics[width=0.5\textwidth]{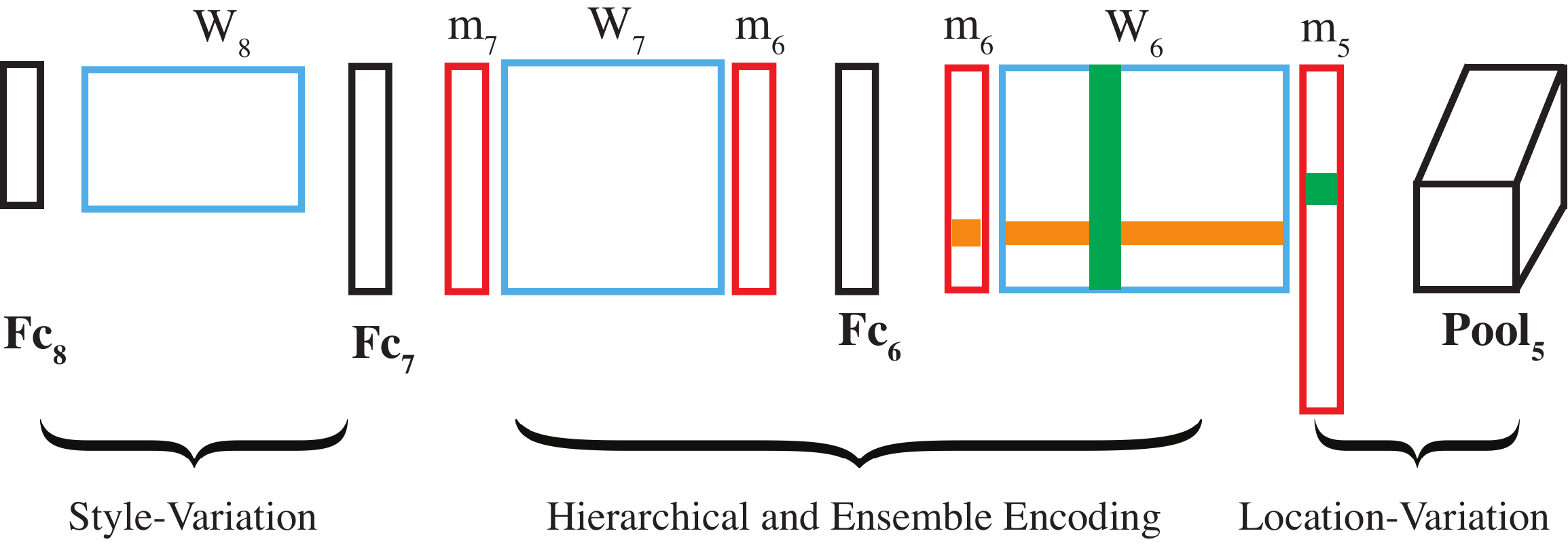}
\caption{Illustration of the organization of Sec.~\ref{sec:model_fc}.
We decompose the fully-connected layers into four components and each subsection explains how the intra-class knowledge is captured and represented.}
\label{fig:fc_pipeline}
\end{figure}
For the class visualization task~\cite{Simonyan14}, notice that the back-propagated gradient from the CNN (data energy) in Eqn.~\ref{Eqn:E_f8} is a series of matrix multiplication.
{\small
\begin{align}
\mbox{fc}_8^t\frac{\partial \mbox{fc}_8}{\partial I}=\mbox{fc}_8^t\frac{\partial \mbox{fc}_8}{\partial \Phi}\frac{\partial \Phi}{\partial I}= W_8^t(m_7W_7^Tm_6W_6^Tm_5)\frac{\partial \Phi}{\partial I},\label{Eqn:f8_grad}
\end{align} 
}

where $W_8^t$ is the $c$-th row of $W_8$,
$m_5,m_6,m_7$ are the relu mask computed on $I$ for fc$_{6-7}$ during feedfoward stage.

Given the learned weights ($W_{6-8}$),
we can turn on/off units from the mask $m_{5-7}$
to sample different structure of the fc layers (``neural pathways'') by multiplying different sub-matrices from the learned weights.
Another view is that the class-specific information is stored in $W_8^t$,
and it can be decoded by different $W_{6-7}$ structures through the relu mask $m_{5-7}$.
\cite{Simonyan14} uses all the weights in $W_{6-7}$,
which leads to a composite template of object parts of different styles in all places (Fig.~\ref{fig:inv_p5}\textcolor{red}{b}).

Below, by controlling the mask $m_{5-7}$,
we show that 
CNN captures two kinds of intra-class knowledge (location and style), which is encoded with an ensemble and hierarchical representation (Fig. ~\ref{fig:fc_pipeline}).

\subsection{Location-variation ($m_5$)}
\label{subsec:fc_loc}
The output from the last convolutional layer is pool$_5$, 
which is semantically shown to be effective as object detectors~\cite{Zhou14}.
Pool$_5$ features (and its relu mask $m_5$) have the 6x6 spatial dimension
and we can visualize an object class within a certain receptive field (RF)
by only opening a subset of spatial dimensions (e.g. $k\times k$ patches) during optimizing Eqn.~\ref{Eqn:E_f8}.

\begin{figure}[ht]
\centering
\includegraphics[width=0.5\textwidth]{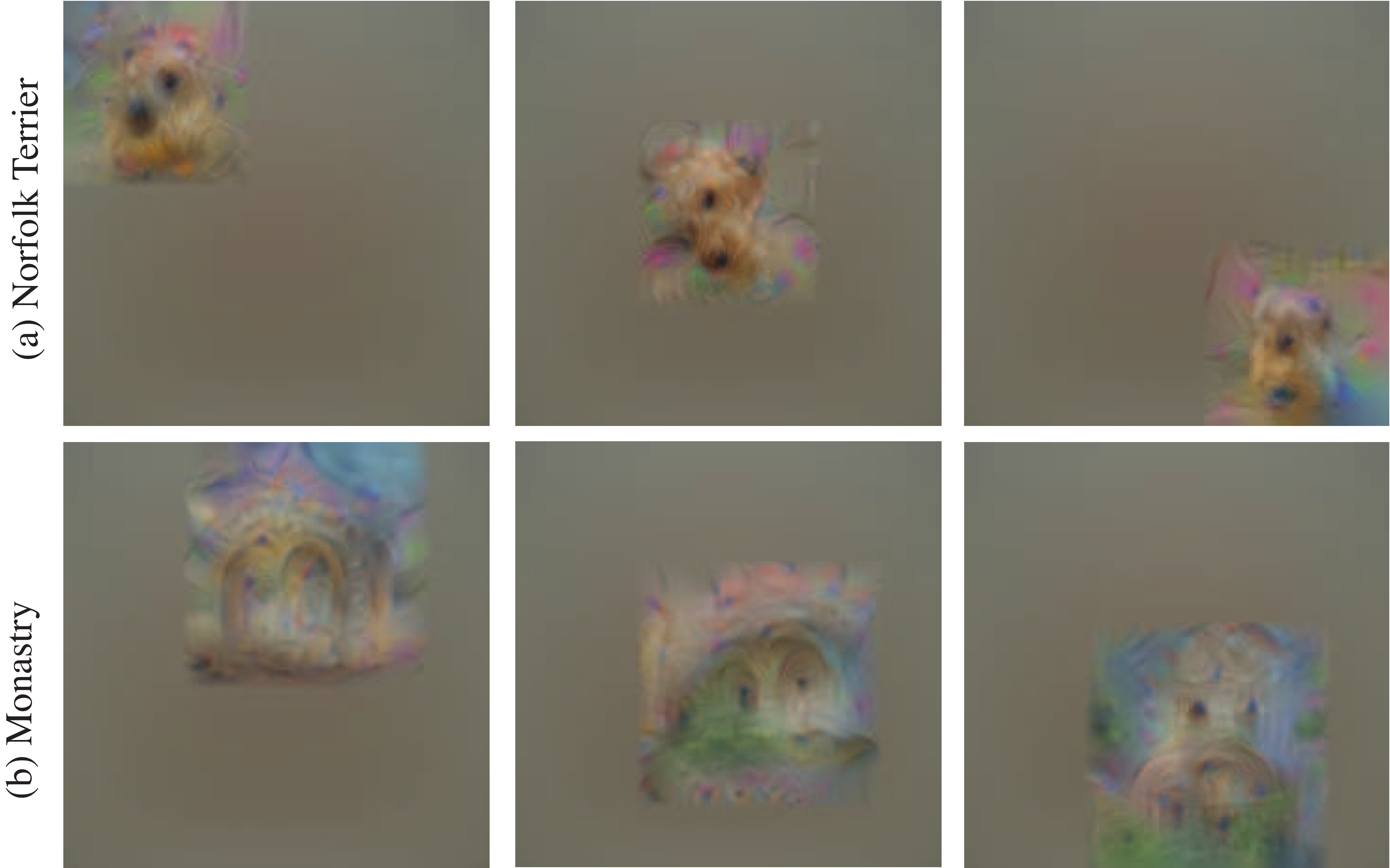}
\caption{Illustration of location-based variation. learn different spatial prior.}
\label{fig:app_p5}
\end{figure}

In Fig.~\ref{fig:app_p5}, we show that the ``terrier'' class doesn't have much variation at each RF, as it learns the dog head uniformly. 
On the other hand, the ``monastery'' class displays heterogeneity, as it learns domes at the top of the image, windows in the middle and doors at the bottom.

\subsection{Content-variation ($m_7$)}
\label{subsec:fc_content}
fc$_7$ has been used as the image semantic space and it has been reported indicative for image retrieval.

\begin{figure*}[t]
\centering
\includegraphics[width=1.0\textwidth]{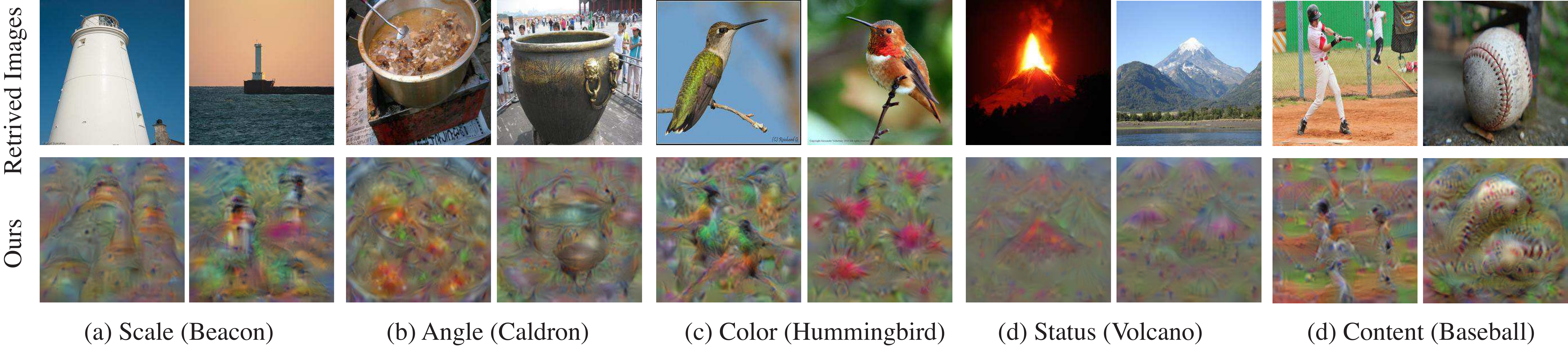}
\caption{Visualization of various topics learned by CNN with retrieved images and our visualization.
These templates capture intra-class variation of an object class: (a) scale, (b) angle, (c) color, (d) status and (e) content.}
\label{fig:fc_f7}
\end{figure*}

\noindent\textbf{Semantic Space as Convex Cone in fc$_7$}
Notice that fc$_8$ is a linear combination of fc$_7$.
Thus, in the fc$_7$ feature space,
if two feature vectors $f_1$ and $f_2$ have the same predicted top-1 class,
then any feature vector $f\in\Omega$ (linear cone) will have the same top-1 prediction
\begin{align}
\Omega = \{\lambda (1-\alpha) f_1 + \alpha f_2): \lambda>0, \alpha\in[0,1]\}\nonumber 
\end{align}
Thus, given the training examples
Namely if two 
linear polytope
(NMF)
In Fig.~\ref{fig:fc_f7}, we show the clusters result of the training examples,
which capture different pose or content of the object, which we calls the ``style'' of the object.

\noindent\textbf{fc$_7$ Topic Visualization}
Given the learned fc$_7$ topic above,
we can apply its relu mask to $m_7$ during optimizing
Eqn.~\ref{Eqn:E_f8}.

\subsection{Ensemble Encoding ($m_{6-7}$)}
\label{subsec:fc_en}
\begin{figure}[h]
\centering
\includegraphics[width=0.45\textwidth]{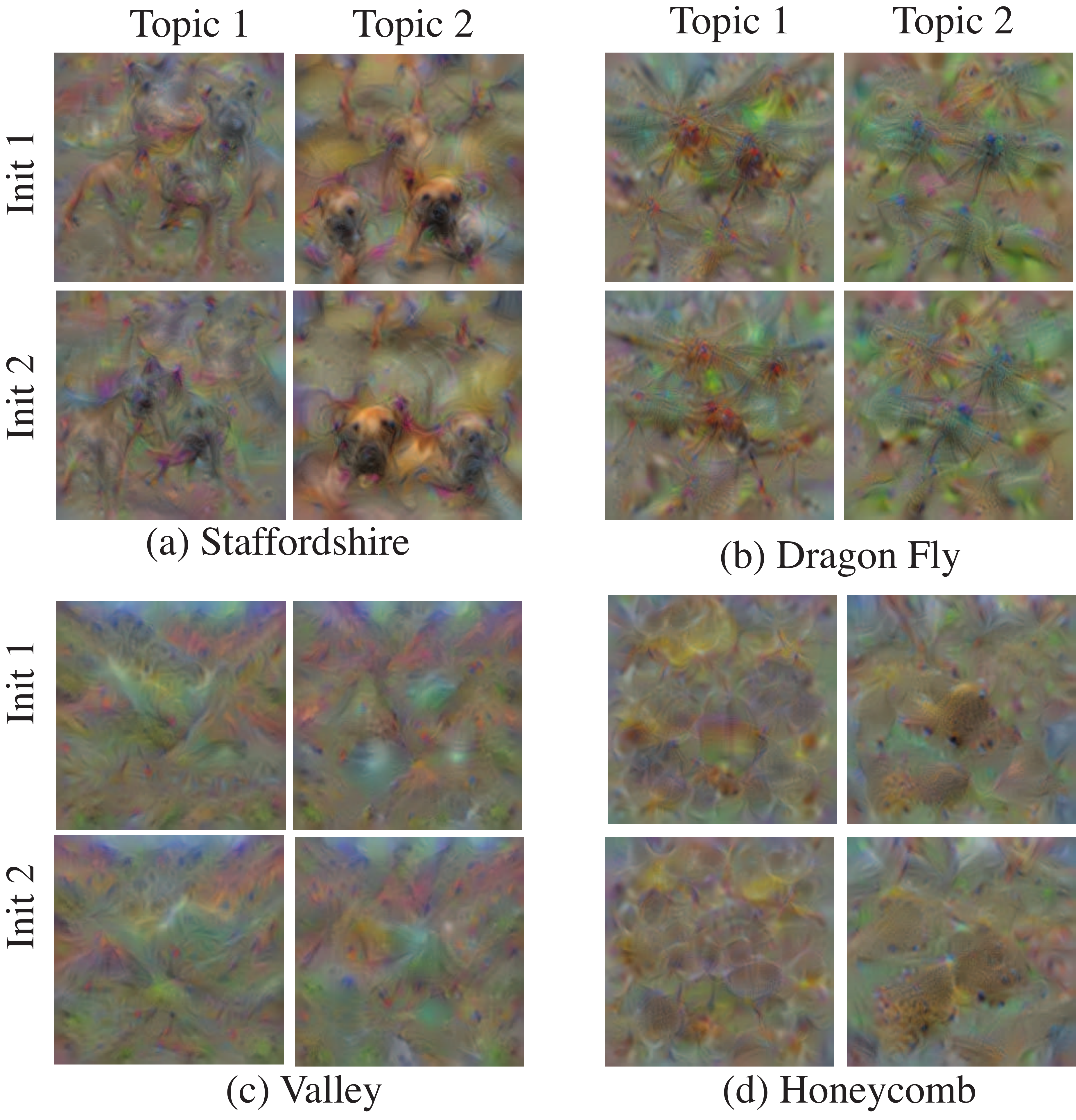}
\caption{Visualization of ensemble encoding.
We show that different dropout model
captures different aspects of a class in terms of
(a) pose,
(b) species,
(c) spatial layout,
and (d) scale
}
\label{fig:fc_f6}
\end{figure}

During training, the dropout trick makes CNN an ensemble model by randomly setting 50\%
of the fc$_{6-7}$ features to be 0,
which is equivalent to turn off half of $m_{6-7}$.
Below, we try to understand what each 
single $m_{6-7}$ model learns
by reconstructing images according to Eqn.~\ref{Eqn:E_f8}.
We randomly sample 2 pairs of $m_{6-7}$ 
correspond to different styles of the objects
and reconstruct the image with 2 different random initialization (Fig.~\ref{fig:fc_f6}).
Interestingly, different models captures different style of the object, 
where the variation across random initialization has smaller effect on the style.

\subsection{Hierarchical Encoding ($m_{5-7}$)}
\label{subsec:fc_hier}
Given an image, we can define its binary code by its relu masks $m_{5-7}$.
~\cite{Agrawal14} discovers that these binary code achieves similar classification result
as their corresponding features.
Similar to dropout model visualization, 
we invert the hash code by 
masking weight matrices $W_{6-7}$ with these binary hash code,
namely constraining CNN to generate images only from these binary masks.
We define three different binary hash code representation for an image
with increasing amount of constraints: $\{m_7\}, \{m_{6-7}\}\{m_{5-7}\}$.
During optimization,
we replace $(W_7,W_6)$
with $(m_7^hW_7,W_6)$, $(m_7^hW_7m_6^h,W_6)$ and $(m_7^hW_7m_6^h,m_6^hW_6m_5^h)$
in Eqn.~\ref{Eqn:E_f8} respectively.

\begin{figure}[h]
\centering
\includegraphics[width=0.45\textwidth]{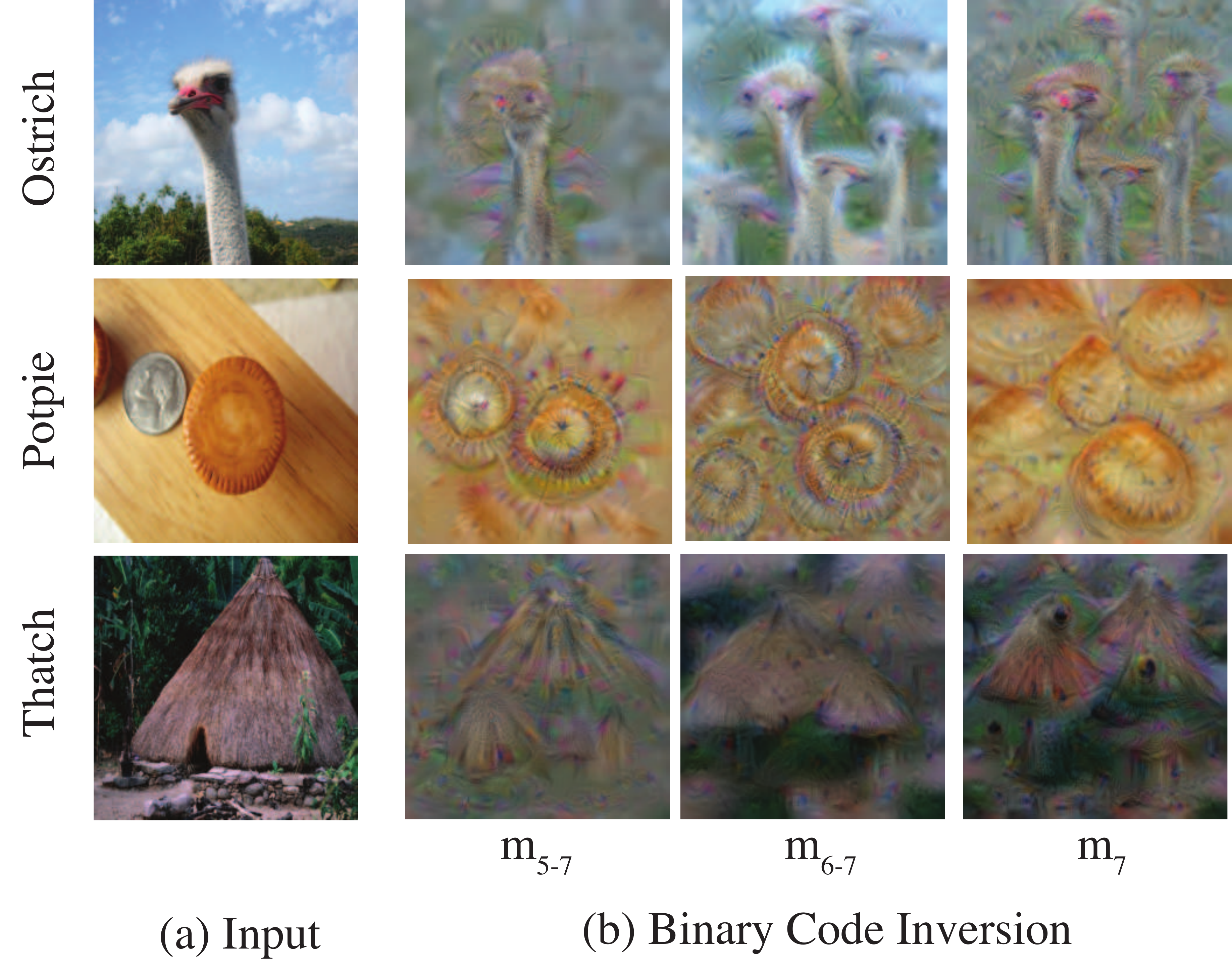}
\caption{Illustration of hierarchical encoding.}
\label{fig:fc_hash}
\end{figure}

\section{Experiments}
\label{sec:result}
\subsection{CNN Visualization Comparison}
\label{subsec:exp_p5}
\begin{figure}[t]
\centering
\includegraphics[width=0.5\textwidth]{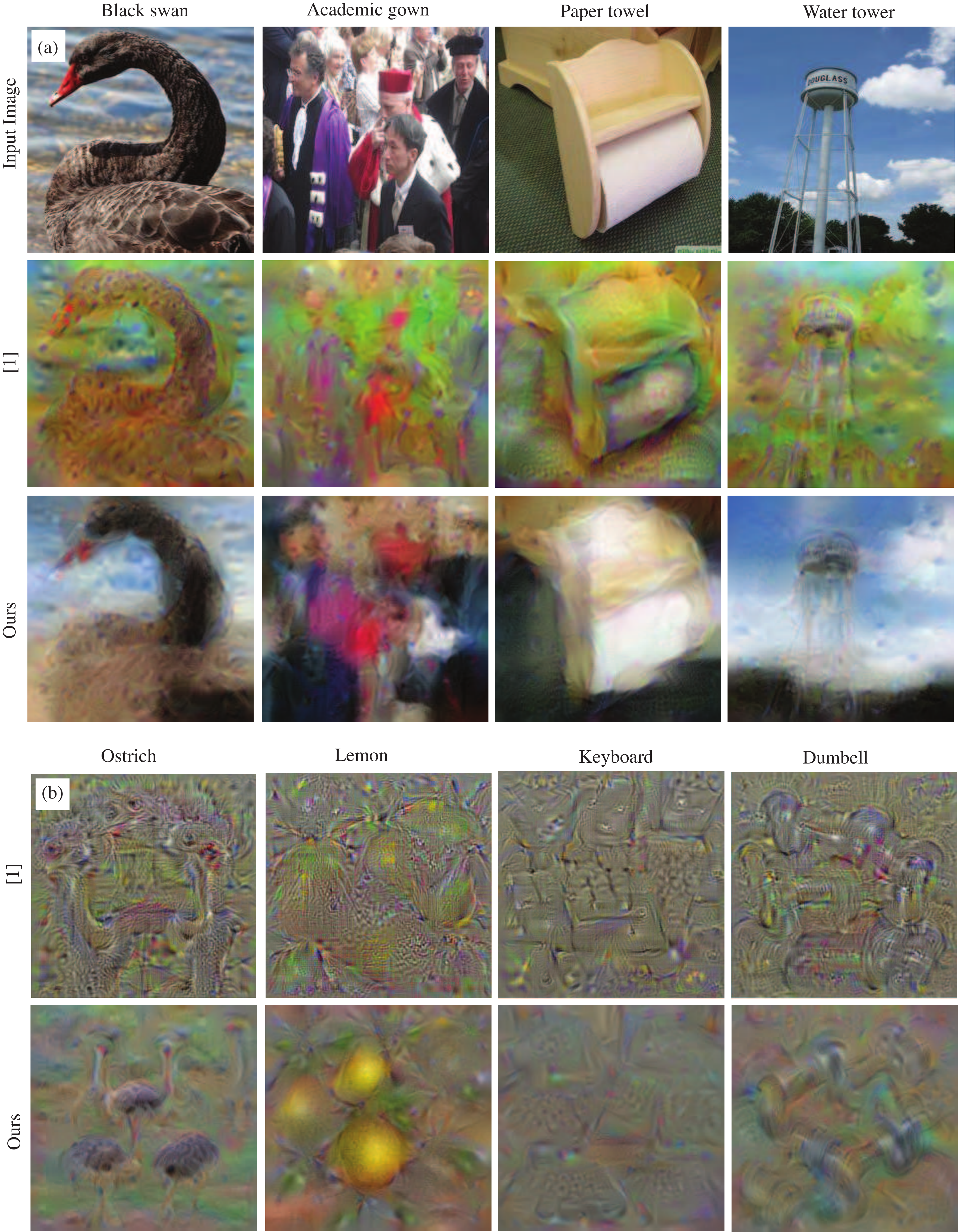}
\caption{Qualitative comparison for CNN visualization.
We compare 
(a) on CNN feature (pool$_5$) inversion  with~\cite{Mahendran15}
and (b) on CNN class visualization  with~\cite{Simonyan14}.}
\label{fig:inv_result}
\end{figure}

For CNN feature inversion, we provide qualitative comparison with 
the previous state-of-the-art~\cite{Mahendran15},
and our results look more natural with the
help of the patch-prior regularization (Fig.~\ref{fig:inv_result}\red{a}).
For quantitative results, 
we collect 100 training images from different classes in the validation set of ImageNet. 
We use the same parameters for both~\cite{Mahendran15} and ours, where the only difference is our patch-prior regularization.
In addition, we empirically found that whiten the image as a pre-procession helps to improve the image regularization without much trade-off for the feature reconstruction error.
For error metric, we use the relative $L_2$ distance between the input image $I_0$ and the reconstructed image $\hat I$ as $\|I_0-\hat I\|_2/\|I_0\|_2$.
We compare our algorithm with two version of~\cite{Mahendran15}: initialized from white noise (~\cite{Mahendran15}+rand) or the same patch database for ours (~\cite{Mahendran15}+~\cite{Long14}).
Shown in Table \ref{table:inv_table},
ours achieves significant improvement.
Notice that,
with the whitening pre-procession and the recommended parameters~\cite{Mahendran15},
most runs have feature reconstruction error $<1\%$,
and we here focus on one whose estimation is closer to the ground truth.

For class visualization task,
as there is no ground truth for sampling images from a given class,
we provide more qualitative results for different kinds of objects (animal, plant and man-made) in Fig.~\ref{fig:inv_result}\red{b}.
Compared to~\cite{Simonyan14},
our visualization results are
closer to natural images
and are easier to interpret.

\begin{table}[ht]
\begin{center}
\begin{tabular}{llll}
\hline
Method
&~\cite{Mahendran15}+rand
&~\cite{Mahendran15}+~\cite{Long14}
&Ours\\
\hline
Error&0.51&0.45&0.32\\
\hline
\end{tabular}
\end{center}
\caption{Quantitative comparison of pool$_5$ feature inversion methods. 
Conditioned on the feature reconstruction error less than a threshold, we compare the distance of the estimated image from the original input. Our method outperform the previous state-of-the-art~\cite{Mahendran15} with two different initializations.}
\label{table:inv_table}
\end{table}

\subsection{Image Completion with Learned Styles}
Given the mask of an image,
we here show the qualitative results on object insertion and modification to explore the potential usage of such object-level knowledge for low-level vision with its top-down semantic understanding of the image.

\begin{figure}[t]
\centering
\includegraphics[width=0.5\textwidth]{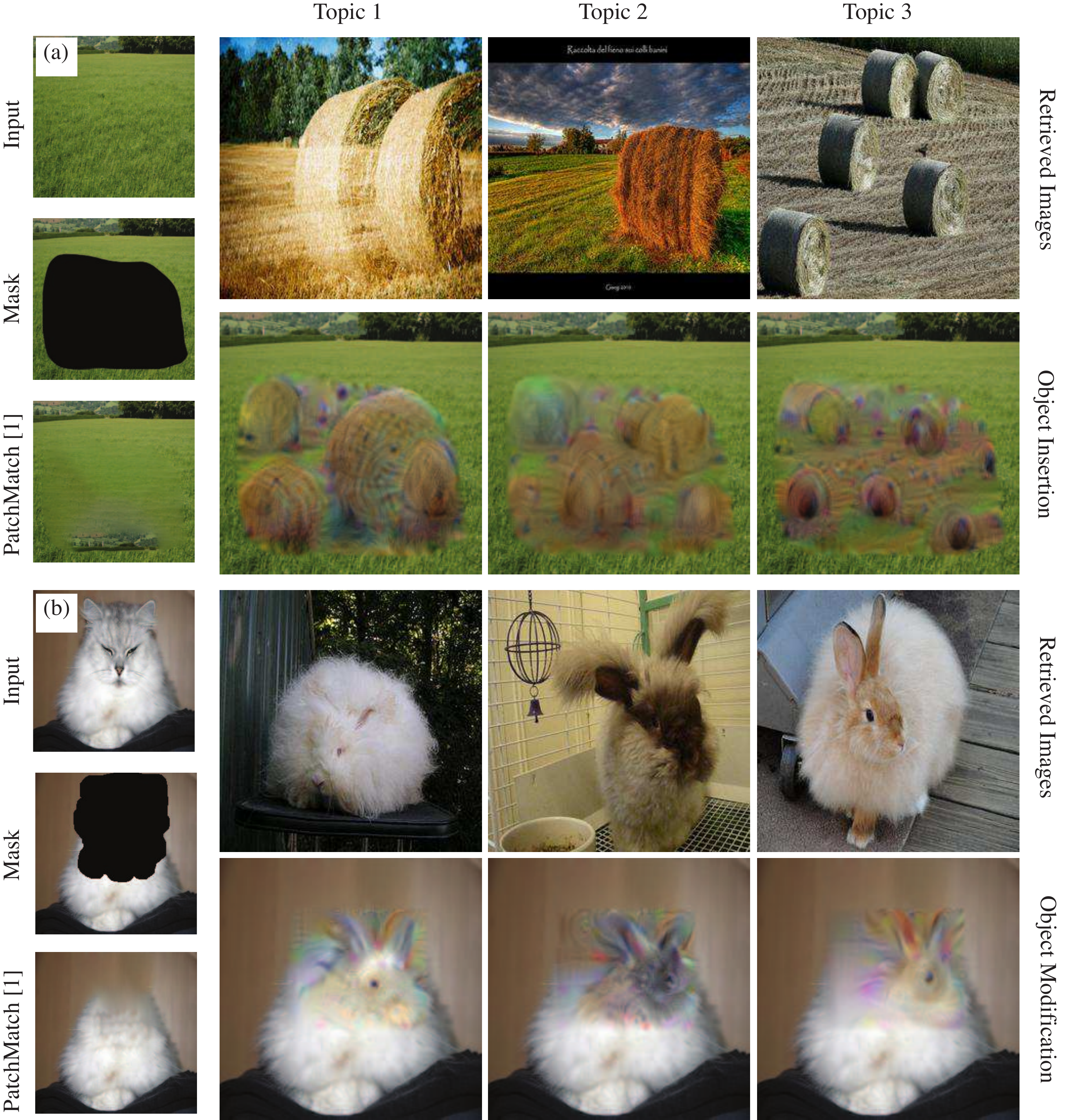}
\caption{Image insertion and modification results. 
Given an image, CNN can not only understand its semantic information to predict the object to insert or the object part to change,
but also fill the mask with 
a specified fc$_7$ style 
using the similar technique as fc$_7$ style visualization.}
\label{fig:app_ip}
\end{figure}

\noindent\textbf{Object insertion from context}
Given a scene image (Fig.~\ref{fig:app_ip}\red{a}),
~\cite{Barnes09} can only fill in grass texture
due to the lack of top-down image understanding.
CNN, on the other hand, can predict relevant object class labels
due to their co-ocurrence in training images.
For this example, the top-1 prediction for the grassland image is ``Hay''. 
Our goal here is to inpaint the hay objects with different styles.

We first follow Sec.~\ref{subsec:fc_content} to learn the styles of hay objects from the Imagenet validation data.
We visualize each topic with a natural image retrieved by it in the top row (Fig.~\ref{fig:app_ip}\red{a}), which correspond to different scales of the hay.
Given a fc$_7$ style topic,
we can insert objects in the image by the procedure similar to our fc$_7$ topic visualization, 
where only pixels inside the mask are updated with the gradient from Eq.~\ref{Eqn:E_f8}.
In the second row, we see different styles of hays are blended with the grassland (Fig.~\ref{fig:app_ip}\red{a}).

\noindent\textbf{Object Modification}
Besides predicting object class based on context information,
CNN can locate the key parts of an object by finding the regions of pixels with high magnitude gradient $\partial \mbox{fc}_8/\partial I$~\cite{Simonyan14}.
Given an input image of a persian cat (Fig.~\ref{fig:app_ip}\red{b}),
we use a simple thresholding and hole filling to find the support of its key part, the head.
Instead of filling the mask with furs as PatchMatch does,
CNN predicts the masked image as ``Angora'' based on the fur information from the body. 
Following the similar procedure as above,
we first find three styles of angoras, 
which correspond to different sub-species with different physical features (e.g. head color), visualized with retrieved images in the third row.
Our object modification result is shown on the bottom row, which change the original persian cat in an interesting way.
Notice that the whole object modification pipeline here is automatic and we only need to specify the style of the angora, as the mask is generated from key object part located by CNN.

\section{Conclusion}
In this work, 
we analyze how CNN model the intra-class variation for each object class in fully-connected layers
through an improved visualization technique.
We find CNN not only captures the location-variation and style-variation,
but also encodes them in a hierarchical and ensemble way.

{\small
\bibliographystyle{ieee}
\bibliography{main_arxiv}
}

\end{document}